\documentclass[sigconf]{acmart}

\AtBeginDocument{%
  \providecommand\BibTeX{{%
    \normalfont B\kern-0.5em{\scshape i\kern-0.25em b}\kern-0.8em\TeX}}}

\setcopyright{acmcopyright}
\copyrightyear{2020}
\acmYear{2020}
\acmDOI{10.1145/XXXXXX.XXXXXX}

\acmConference[KDD '20] {26th ACM SIGKDD Conference on Knowledge Discovery and Data Mining}{August 23--27, 2020}{Virtual Event, USA}
\acmBooktitle{26th ACM SIGKDD Conference on Knowledge Discovery and Data Mining (KDD '20), August 23--27, 2020, Virtual Event, USA}
\acmPrice{15.00}
\acmISBN{978-1-4503-7998-4/20/08}

\usepackage{bm}
\usepackage{amsmath}
\usepackage{threeparttable}
\usepackage{color}
\usepackage{multirow}
\usepackage{bm}

\usepackage{fancyhdr}

\usepackage{subfig}

\usepackage{color}

\newcommand{\eg}{{e.g.}}
\newcommand{\ie}{{i.e.}}

\newcommand{\etc}{\emph{etc}}

\newcommand\blfootnote[1]{%
  \begingroup
  \renewcommand\thefootnote{}\footnote{#1}%
  \addtocounter{footnote}{-1}%
  \endgroup
}

\begin{document}
\fancyhead{}

\title{General-Purpose User Embeddings based on Mobile App Usage}


\author{Junqi Zhang$^{1*}$, Bing Bai$^{1*}$, Ye Lin$^1$, Jian Liang$^1$, Kun Bai$^1$, Fei Wang$^2$}
\affiliation{%
  \institution{$^1$Cloud and Smart Industries Group, Tencent, China\\$^2$Cornell University, USA}
}
\email{{benjqzhang, icebai, yessicalin, joshualiang, kunbai}@tencent.com}
\email{few2001@med.cornell.edu}

\renewcommand{\shortauthors}{Junqi Zhang, Bing Bai, Ye Lin, Jian Liang, Kun Bai, Fei Wang}

\begin{abstract}
In this paper, we report our recent practice at Tencent for user modeling based on mobile app usage. User behaviors on mobile app usage, including retention, installation, and uninstallation, can be a good indicator for both long-term and short-term interests of users. For example, if a user installs \emph{Snapseed} recently, she might have a growing interest in photographing. Such information is valuable for numerous downstream applications, including advertising, recommendations, \etc. Traditionally, user modeling from mobile app usage heavily relies on handcrafted feature engineering, which requires onerous human work for different downstream applications, and could be sub-optimal without domain experts. However, automatic user modeling based on mobile app usage faces unique challenges, including (1)~retention, installation, and uninstallation are heterogeneous but need to be modeled collectively, (2)~user behaviors are distributed unevenly over time, and (3)~many long-tailed apps suffer from serious sparsity. In this paper, we present a tailored AutoEncoder-coupled Transformer Network~(AETN), by which we overcome these challenges and achieve the goals of reducing manual efforts and boosting performance. We have deployed the model at Tencent, and both online/offline experiments from multiple domains of downstream applications have demonstrated the effectiveness of the output user embeddings. \blfootnote{$*$ Equal contributions from both authors. This work is done when Junqi Zhang works as an intern at Tencent.}
\end{abstract}

\ccsdesc[500]{Information systems~Data mining}

\keywords{user modeling; embeddings; autoencoder; transformer; app usage}

\maketitle

\section{Introduction}
\label{sec:introduction}

Personalized mobile business, \eg, recommendations, and advertising, often require effective user representations. For better performance, user modeling in industrial applications often considers as much information as possible, including but not limited to gender, location, interested tags, accounts subscribed, and shopping interests~\cite{liu2019real-time}.
Among which, user behaviors on mobile app usage, including~\emph{retention}~(which apps are currently installed on the phone), \emph{installation}~(when and which apps were ever installed recently), and \emph{uninstallation}~(when and which apps were removed from the phone recently), contain rich information about both long-term and short-term user interests~\cite{lu2014mining}.
For example, if a user installs \emph{Google Photos}, \emph{Snapseed}, and \emph{Instagram}, there is a good chance that she is an enthusiast of mobile photographing. If a user installs the popular game \emph{Honor of Kings, a.k.a. Arena of Valor} recently, she might be a new gamer and is wondering how to play better. Such information is valuable for various downstream applications, and how to utilize them better is an exciting problem worthy of solving.

Traditionally, mining from mobile app usage relies on task-specific handcrafted features. For example, recommending a new game app to users who have installed similar games can help avoid recommending to non-gamers. However, handcrafted feature engineering often requires substantial human efforts, and maybe sub-optimal when domain experts are absent. To improve efficiency and effectiveness, an automatic generation for general-purpose user representations from user behaviors on mobile app usage is in need.

We have been working towards this goal since mid 2019, and several versions of models have been deployed. In this paper, we outline the most recent practice at Tencent. The main challenges of building general-purpose user representations for multiple downstream applications include:

\begin{figure}[t]
  \centering
  \includegraphics[width=0.975\linewidth]{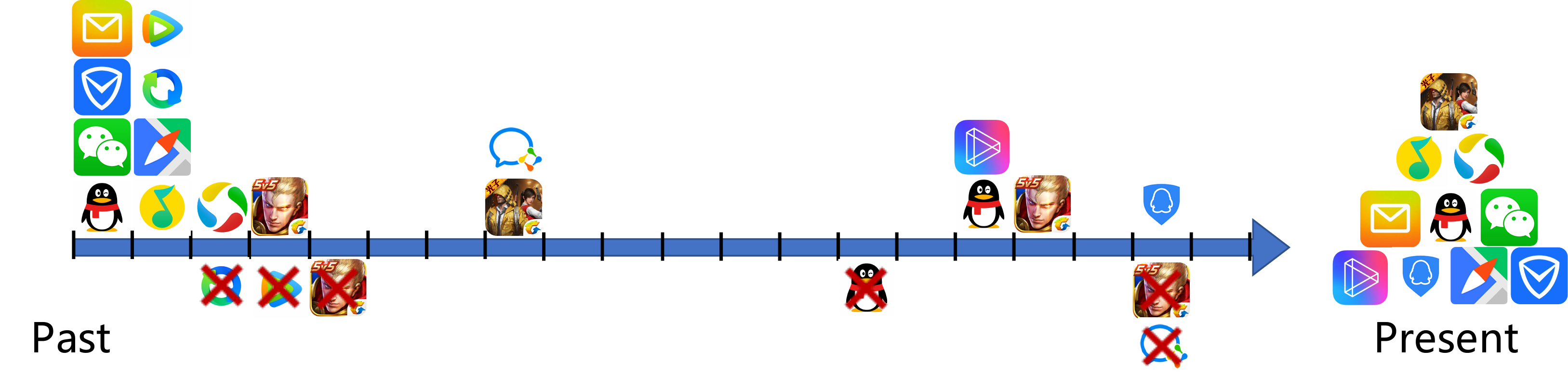}
  \caption{Illustration of retention, installation, and uninstallation. Operations of (un)installation are low-frequency and unevenly distributed over time.}
  \label{fig:behaviors}
\end{figure}

\begin{figure*}[th]
  \centering
  \includegraphics[width=0.975\linewidth]{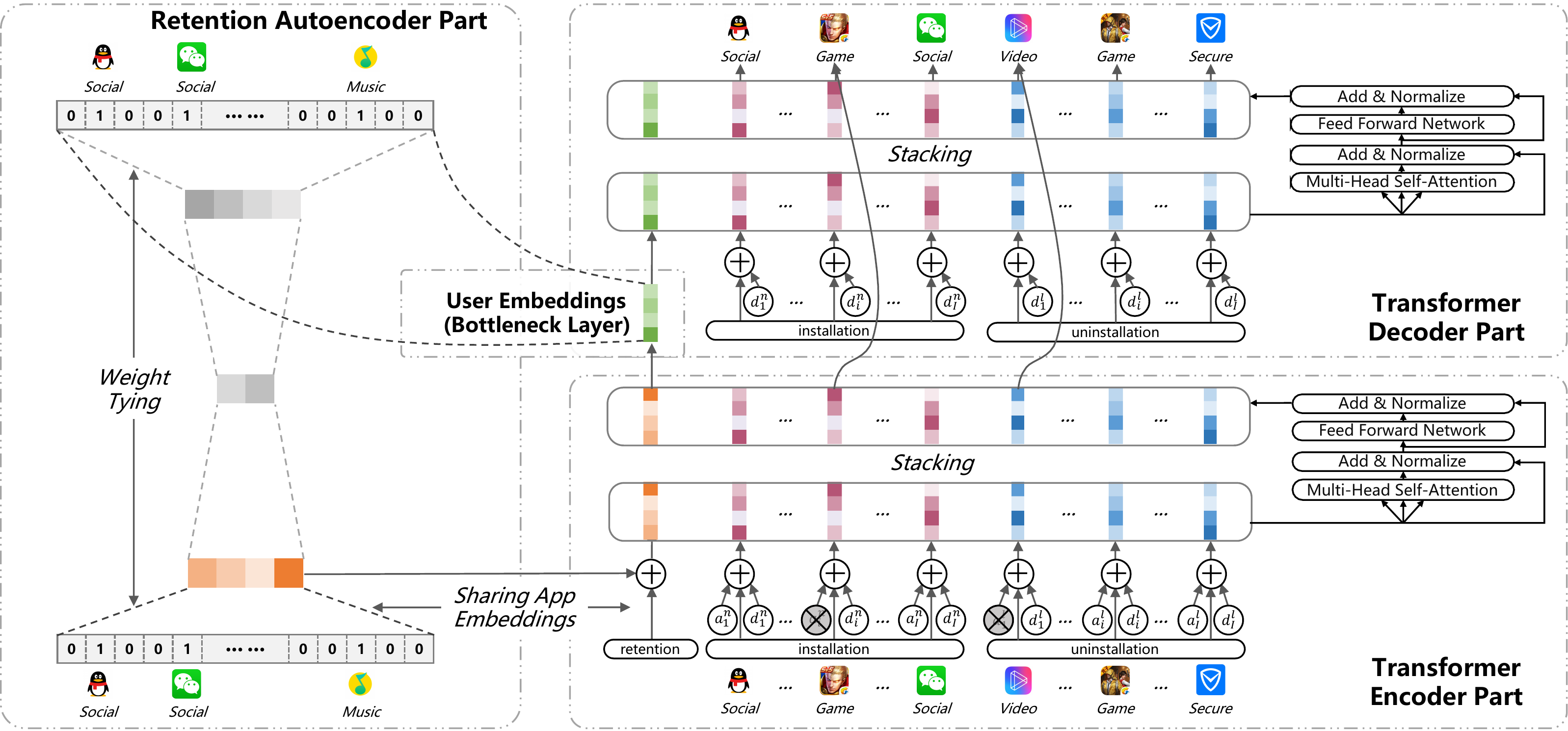}
  \caption{The overview of the proposed AETN model. The model is tailored for learning effective user embeddings from mobile app usage unsupervisedly. The retention autoencoder part aims to learn good representations for apps and user retention based on co-occurrence relationships. The transformer encoder part models the retention, installation, and uninstallation information collectively, and maps the user to an embedding at the bottleneck layer. The user embeddings are forced to maintain as most information as possible by the transformer decoder part, which reconstructs the installation and uninstallation sequences. Besides, the user embeddings also need to be able to reconstruct the retention.}
  \label{fig:system}
\end{figure*}

\begin{itemize}
    \item{\bfseries Retention, installation and uninstallation need to be modeled collectively.} They represent the preference of users from different aspects, and building representations for the three parts separately and then concatenating them may limit the performance. For example, for users who have installed multiple games, uninstalling a game app may only indicate that she has finished the game and wants to start a new one. While for a user who has not installed other games, immediately uninstalling after installation may suggest that she does not like this kind of game at all. Modeling such complex relationships using traditional recurrent neural networks~(RNNs) is challenging.

    \item{\bfseries Actions of (un)installing apps are low-frequency and unevenly distributed over time.} Figure~\ref{fig:behaviors} presents a demo of app installation and uninstallation records of a user. As excitement over the new phone fades, most users only install or uninstall apps when they need to. Moreover, users usually do not operate for even a month but may suddenly install or uninstall several apps in a single day. In this case, various intervals between every two behaviors are not omittable. Although RNN-based models have succeeded in analyzing user activities~\cite{hidasi2015session,li2017neural}, the behaviors in those scenarios are usually with notably higher-frequency and nearly even distribution over time. Therefore, traditional RNNs may not perform well for this task.
    
    \item{\bfseries Many long-tailed apps suffer from serious sparsity.} Popular apps like \emph{Wechat} and \emph{Alipay} have been installed on almost all the smartphones in China, while long-tailed apps may only have a few hundreds of installations among one million users. However, user's behaviors over the long-tailed apps often reflect one's personalized interests better. Building effective user representations need to utilize the information from long-tailed apps without suffering from severe sparsity.
\end{itemize}

To achieve the goal, we design a tailored AutoEncoder-coupled Transformer Network~(AETN) to analyze user behaviors on mobile app usage. The model follows a classical encoder-decoder framework with a bottleneck for user representation learning, and utilizes a multi-objective joint training scheme for parameter learning. Figure~\ref{fig:system} shows the general framework. The model mainly consists of three parts, \ie, the retention autoencoder part, the (stacked) transformer encoder part, and the (stacked) transformer decoder part. The three parts are tied through parameter sharing and trained jointly. The proposed model is entirely unsupervised and carefully optimized for learning user embeddings from mobile app usage.

The retention autoencoder serves as a foundational part of AETN. From the co-occurrence relationship of apps in retention data, it learns and shares effective app embeddings with the transformer network. As one of the designs to alleviate the problem of sparsity, we model the embeddings of apps with both app IDs and their corresponding category IDs. Therefore, if the usage of an app is gravely sparse, at least the category ID can provide some information. Another design is weight tying between the encoder and the decoder. Note that we only tie the first and the last layer of the autoencoder to leave enough flexibility. Weight tying can significantly reduce the number of free parameters, and hasten the convergence~\cite{hidasi2018recurrent}. Together with app embeddings, effective representations of user retention are obtained and provided to the transformer parts.

On the other hand, the transformer parts model the retention, installation, and uninstallation collectively, and output the final user embeddings. Transformer networks have been proved effective for modeling (multiple) sequences and obtaining contextual representations in natural language processing~\cite{devlin2019bert}. Inspired by BERT~\cite{devlin2019bert}, in this paper, we use a stacked transformer network to consolidate different types of information.

The transformer encoder part receives the user retention, shared app embeddings, date embeddings, and behavior type embeddings~(retention, installation, and uninstallation) as input. Thus, the inputs altogether include complete information on when and whether users install or uninstall what apps as well as their current status of app usage. The date embeddings make the transformer suitable for modeling user behaviors that are low-frequency and distribute unevenly over time. Besides, we also introduce a \emph{masked app prediction} task like BERT~\cite{devlin2019bert} to help extract information more productively.

After compressing all the input information at a bottleneck layer, the (stacked) transformer decoder part tries to reconstruct the (un)installation sequences. The reconstruction follows a manner similar to non-autoregressive translation~\cite{gu2017non}. And the date embeddings, as well as the behavior type embeddings, are used as the queries. We also reconstruct the retention data from the bottleneck layer with a multi-layer perceptron network. The reconstruction processes force the bottleneck to retain as much information as possible from the original input through the transformer encoder.

Besides, we use weight tying in the output layers of both the transformer encoder and the decoder.
Moreover, to better encourage information interaction within the transformer network, we proposed a modified multi-head self-attention mechanism where the representations of retention or bottleneck are fed to attention mechanisms more directly during every attention step.
All the components mentioned above are trained jointly over data from tens of millions of users of Tencent. Representations from the bottleneck of the transformer network are used as general-purpose user embeddings, which can fertilize many downstream applications that require user representations. The main contributions of our work are summarized as follows:
\begin{itemize}
    \item We introduce our recent practice of general-purpose user embedding learning based on mobile app usage for multiple downstream applications. 
    
    \item We propose a tailored model AETN to achieve the goal. With a carefully-designed neural network structure, the autoencoder-coupled transformer network overcomes the serious sparsity of long-tailed apps and the uneven distribution of activities, and models user behaviors on mobile app usage collectively. Our code is publicly available.\footnote{https://github.com/Junqi-Zhang/AETN\label{code}}
    
    \item The cost of model training is acceptable in real application scenarios. Extensive online and offline experiments verify the effectiveness of the proposed model, which has been deployed in a real system at Tencent and achieved boosted performance in daily business.
\end{itemize}

The rest of the paper is organized as follows. We introduce the background in Section~\ref{sec:background}. Section~\ref{sec:system} and Section~\ref{sec:AETN} describe our high-level system and the detailed design of AETN respectively. We present offline experiments and the online A/B testing in Section~\ref{sec:experiment} and Section~\ref{sec:online}. The details of model deployment are presented in Section~\ref{sec:deploy}. Related work is discussed in Section~\ref{sec:relatedwork}, and Section~\ref{sec:conclusion} draws the conclusion.

\section{Background}
\label{sec:background}
\emph{Tencent Mobile Manager} is currently the most prevalent mobile security and management app in China, which serves nearly one billion users. We provide various auxiliary functionalities, including news recommendations, short video recommendations, app recommendations, \etc. For example, users can reach personalized content feeds, including news, articles, and short videos, from the ``Good Morning'' tab of \emph{Tencent Mobile Manager}, as well as from the ``Find'' tab of \emph{Tencent Wi-Fi Manager}, a wingman app of \emph{Tencent Mobile Manager}.

We have built a data center to support various downstream applications. Traditional handcrafted feature engineering and shallow models may not maximize the value of data, therefore, in terms of user behaviors on mobile app usage, general-purpose user representations are desired. 

\section{System Overview}
\label{sec:system}

In this section, we introduce our AETN-based system from a high-level perspective and review its data processing, model training, and serving components.

\subsection{Data Preprocessing}
\label{sec:preprocessing}
We need to preprocess the user data into a format suitable for subsequent models to handle and also reduce the noise in data. After data preprocessing, each user is represented with one's ``retention'' and four sequences. ``Retention'' is a set of apps installed on one's phone at present. Two of the sequences, representing recent ``installation'' operations, are composed of installed apps and corresponding dates. The rest two sequences represent recent ``uninstallation'' operations. To reduce the noise in user behaviors, we keep the most recent 10 installation or uninstallation operations in a week for each user.

We use the following criteria to select the apps considered in the model.
\begin{itemize}
    \item We manually exclude some top-ranked apps which have been installed on almost every smartphone and can hardly represent user interests, such as \emph{Wechat}. Meanwhile, we keep apps like \emph{Honor of Kings} despite that they are popular, for they could still represent users' personalized interests.
    \item We exclude the apps pre-installed on smartphones by the manufacturers.
    \item We also exclude the niche apps with installed capacities under a threshold.
\end{itemize}

Besides, one app may have multiple \textit{package\_names} for different brands and models of smartphones. They are all merged to avoid duplication. For the categories of apps, we consider relatively finer-grained app categories, for example, we distinguish different subcategories of ``Game'' apps.

\subsection{Model Training and Serving}

After preprocessing data, we train the model and generate the user embeddings with the following steps,
\begin{itemize}
    \item Step 1: Model Training. We train the AETN using tens of millions of users.
    \item Step 2: Inference. We extract user embeddings for all the users, and push the embeddings to a DCache system\footnote{https://github.com/Tencent/DCache} for serving.
    \item Step 3: Serving. Downstream applications can retrieval user embeddings using the feature ID and user IDs. Gradient boosting decision trees~(GBDTs) and neutral networks~(NNs) are typically used as downstream models.
\end{itemize}

More details about the deployment are in Section~\ref{sec:deploy}.

\section{AutoEncoder-coupled Transformer Network}
\label{sec:AETN}
In this section, we first define the notations of user behaviors, followed by the detailed structure of the proposed network. Then, we elaborate on our designs for alleviating the problem of sparsity and our modification to vanilla transformers. Finally, we present the multi-objective joint training scheme for model optimization.

\subsection{Notations of User Behaviors}
\label{sec:notation}

As stated in Section~\ref{sec:preprocessing}, behaviors of each user are preprocessed into one's ``retention'' and four sequences defined as follows.

{\bfseries Retention.} The retention of user $u$ can be represented by a multi-hot vector $\bm{x}_u \in \mathbb{R}^M$, and $x_{um} = 1$ when app $m$ is installed, where $M$ is the number of considered apps.

{\bfseries Installation and Uninstallation.} The four sequences, representing user $u$'s latest $I$ operations on installing or uninstalling apps, are denoted by $\mathcal{S}_u$:
\begin{align*}
\mathcal{S}_u=\big\{&[a^{n}_{u,1},\dots,a^{n}_{u,i},\dots,a^{n}_{u,I}],[d^{n}_{u,1},\dots,d^{n}_{u,i},\dots,d^{n}_{u,I}],\\
&[a^{l}_{u,1},\dots,a^{l}_{u,i},\dots,a^{l}_{u,I}],[d^{l}_{u,1},\dots,d^{l}_{u,i},\dots,d^{l}_{u,I}]\big\}.
\end{align*}
Specifically, $a^n_{u,i}$ represents the ID of $i$-th newly installed app of~$u$, and~$d^n_{u,i}$ is the corresponding date.~$a^l_{u,i}$ and~$d^l_{u,i}$ are the  counterparts for uninstallation. Additionally, $1 \leq a^n_{u,i}$, $a^l_{u,i} \leq M$, and all the operations happened in the latest $T$ time intervals.

Note that in the rest of the paper, we omit the subscript $u$ indicating a user in most notations for simplification.

\subsection{Network Structure}
\label{sec:structure}

As shown in Figure~\ref{fig:system}, an autoencoder for retention, a transformer encoder, and a transformer decoder are three main parts in the proposed model. We connect the latter two parts with a bottleneck layer. There is also an embedding layer for the transformer encoder and another one for the decoder. Details about the network structure are as follows.

{\bfseries Retention Autoencoder.} The AETN employs an autoencoder of three hidden layers to reconstruct and encode one's retention. The autoencoder can be described with triple tuples~$(f^{(p)}, \mathbf{W}^{(p)}, \bm{b}^{(p)})$, where $p\in\{1,2,3,4\}$.~$\mathbf{W}^{(p)}$ and $\bm{b}^{(p)}$ are weights and biases of the \mbox{$p$-th} layer, and~$f^{(p)}$ represents the corresponding activation function. We choose the commonly used \emph{LeakyReLU}~\cite{maas2013rectifier} as the activation function for the former three layers, and $f^{(4)}$ is the \emph{sigmoid} function. Let~$\bm{x}^{(p)}$ represent the outputs of each layer, and it can be calculated as follows:
\begin{equation}
  \bm{x}^{(p)}=f^{(p)}(\bm{x}^{(p-1)}\mathbf{W}^{(p)}+\bm{b}^{(p)}),~p\in\{1,2,3,4\},
\end{equation}
where $\bm{x}^{(0)}$ is normalized from one's retention $\bm{x}$ using the $\ell 2$ norm.

The role of this autoencoder is two-folds. Firstly, it helps to learn high-quality app embeddings from the co-occurrence relationship of apps. The weight matrix of the first hidden layer~$\mathbf{W}^{(1)}$ acts as the shared app embedding matrix~$\mathbf{W}^a$ for the whole network, \ie, we have
\begin{equation}
\mathbf{W}^a=\mathbf{W}^{(1)}\in\mathbb{R}^{M\times{d_{model}}}.
\end{equation}
To further alleviate the problem of sparsity, the shared app embedding matrix is carefully designed and tied with some other weight matrices. More details are provided in Section~\ref{sec:weights}.

Secondly, this autoencoder provides effective representations of user retention for the transformer part. The transformer encoder part needs to be fed with the retention for compressing long-term interests into user embeddings. However, retention is originally in the form of high-dimensional sparse features. This autoencoder encodes retention into the first hidden layer~$\bm{x}^{(1)}\in\mathbb{R}^{d_{model}}$. As a low-dimensional dense encoding,~$\bm{x}^{(1)}$ plays an important role in the transformer encoder part.

\begin{figure}[ht]
  \centering
  \includegraphics[width=0.875\linewidth]{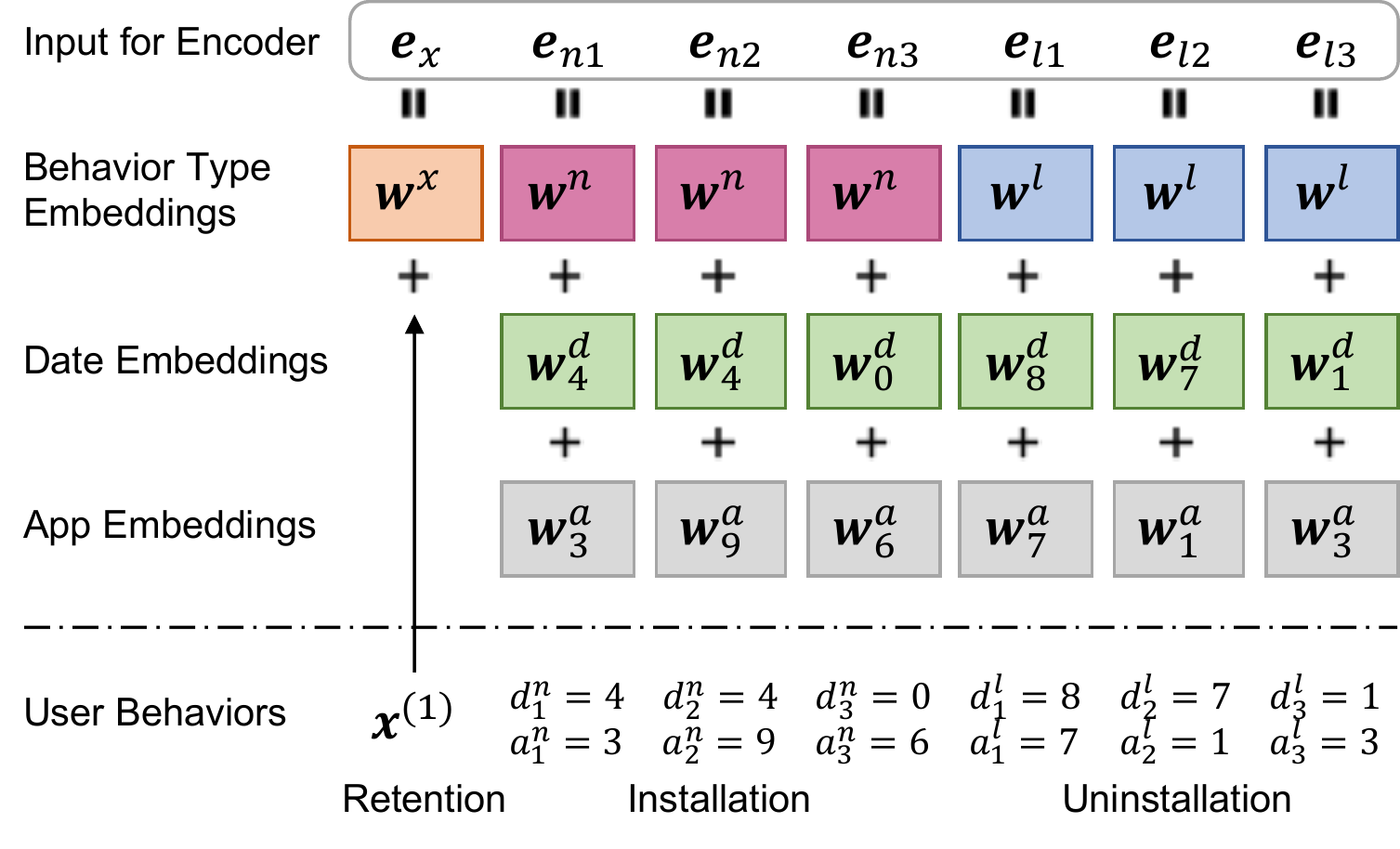}
  \caption{Embedding Layer for Transformer Encoder.}
  \Description{Embedding Layer for Transformer Encoder.}
  \label{fig:encoder_embedding}
\end{figure}

{\bfseries Transformer Encoder \& Its Embedding Layer.} The transformer encoder is the core part of AETN to combine and compress all the information, which does not work without a suitable embedding layer. Inspired by positional encodings~\cite{vaswani2017attention}, we design an embedding layer for the transformer encoder based on the shared app embeddings, date embeddings, and behavior type embeddings, as illustrated in Figure~\ref{fig:encoder_embedding}.

The date embeddings are the key to making the whole network suitable for modeling user behaviors that are low-frequency and unevenly distributed over time. Through date embeddings, the subsequent transformer encoder directly receives the information about when the behaviors happened rather than inferring it from the order of behaviors. We denote the date embedding matrix as $\mathbf{W}^d\in\mathbb{R}^{T\times d_{model}}$, and date $t$ is represented by $\bm{w}^d_t\in\mathbb{R}^{d_{model}}$.

The behavior type embeddings help the model to distinguish different types of user behaviors when integrating them all. For the three user behavior types~(retention, installation, and uninstallation), the embeddings are~$\bm{w}^x,\bm{w}^n,\bm{w}^l\in\mathbb{R}^{d_{model}}$.

Through this embedding layer, we construct the input representations for the transformer encoder, and the input includes complete information about one's mobile app usage.

Our encoder blocks share a similar basic structure with the origin transformer encoder~\cite{vaswani2017attention}, and to encourage the information interaction among different types of behaviors, we make small modifications to the multi-head self-attention mechanism. More details are presented in Section~\ref{sec:self-attention}.
To better extract information from user behaviors, inspired by the \emph{masked language model} task in BERT~\cite{devlin2019bert}, we apply a \emph{masked app prediction} task to installation and uninstallation sequences. The weight matrix of the output \emph{softmax} is denoted by $\mathbf{W}^{\Omega}\in\mathbb{R}^{d_{model}\times M}$. More details about this training task are provided in Section~\ref{sec:train_scheme}.

{\bfseries Bottleneck Layer.} The bottleneck layer is where~(low dimensional) user embeddings, denoted as~$\bm{\widetilde{e}}$, are generated. As the encoder and the decoder fuse in this layer, the compressed information from original inputs becomes the source of information for reconstruction tasks.

On top of the final hidden vector~$\bm{e}^{\Omega}_x$, \ie, the representations corresponding to the retention output by the transformer encoder, we use a single hidden layer autoencoder to further reduce the dimension from $d_{model}$ to $d_{emb}$. The activation function for the bottleneck is \emph{tanh}. Then the reconstructed input of this autoencoder is fed to the transformer decoder part.

In the training scheme, we reconstruct one's retention from her user embedding with a multi-layer perceptron network and the \emph{sigmoid} activation function. The weight matrix of the output layer is denoted as $\mathbf{W}^{\Theta}\in\mathbb{R}^{d_{model}\times M}$.

\begin{figure}[t]
  \centering
  \includegraphics[width=0.875\linewidth]{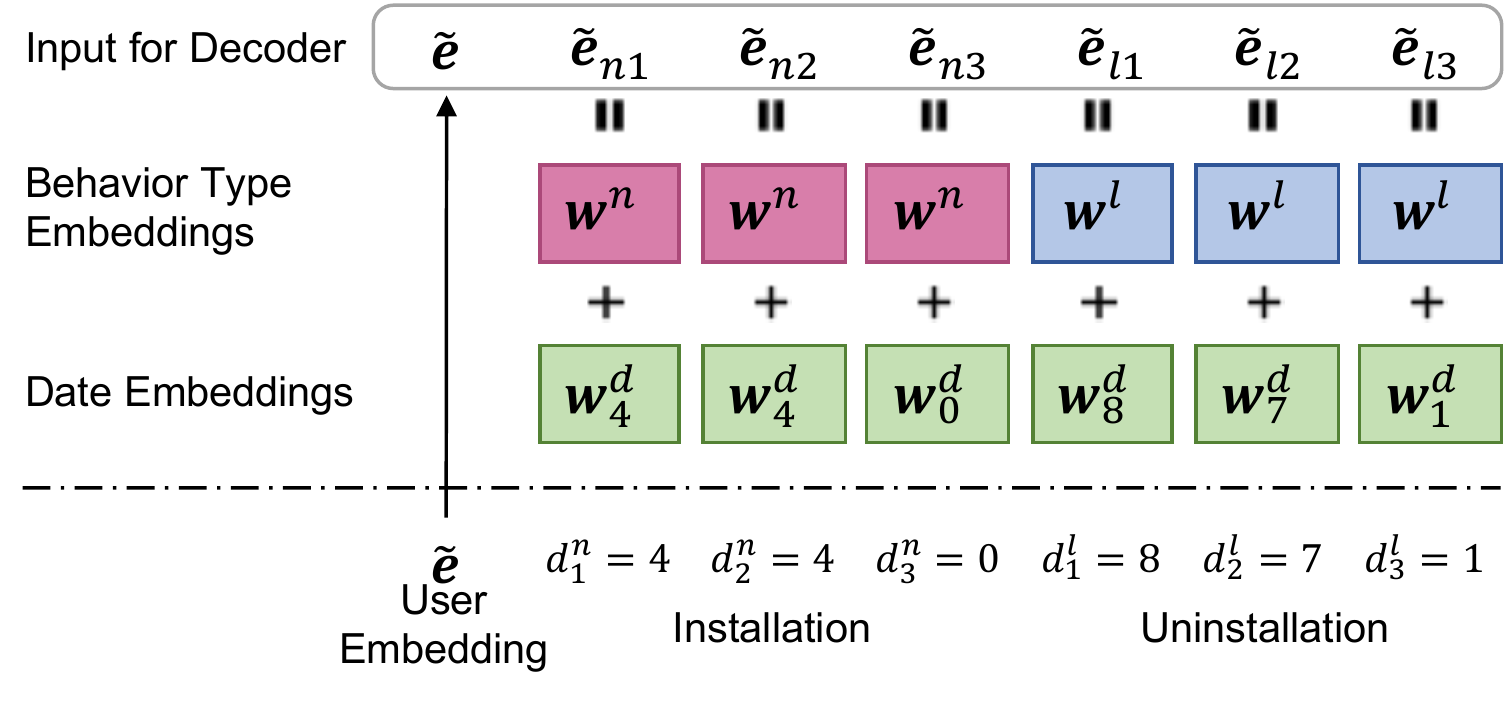}
  \caption{Embedding Layer for Transformer Decoder.}
  \Description{Embedding Layer for Transformer Decoder.}
  \label{fig:decoder_embedding}
\end{figure}

{\bfseries Transformer Decoder \& Its Embedding Layer.} The transformer decoder serves our purpose of reconstructing installation and uninstallation in a non-autoregressive manner~\cite{gu2017non}. 
More concretely, we use the date and the behavior type as queries to search for valuable information from the user embedding to reconstruct corresponding installed or uninstalled apps. For this purpose, we design a new embedding layer for the transformer decode, sharing date embeddings and behavior type embeddings with the embedding layer of the encoder. Figure~\ref{fig:decoder_embedding} shows the details of this embedding layer and the input for the decoder.

To accomplish the task of reconstructing entire installation and uninstallation sequences, we feed all hidden vectors, which is corresponding to the installation and uninstallation, of this decoder into an output \emph{softmax} layer. The weight matrix of this layer is denoted as $\mathbf{W}^{\Phi}\in\mathbb{R}^{d_{model}\times M}$.

\subsection{Weight Matrix Settings}
\label{sec:weights}
\begin{figure}[t]
  \centering
  \includegraphics[width=0.875\linewidth]{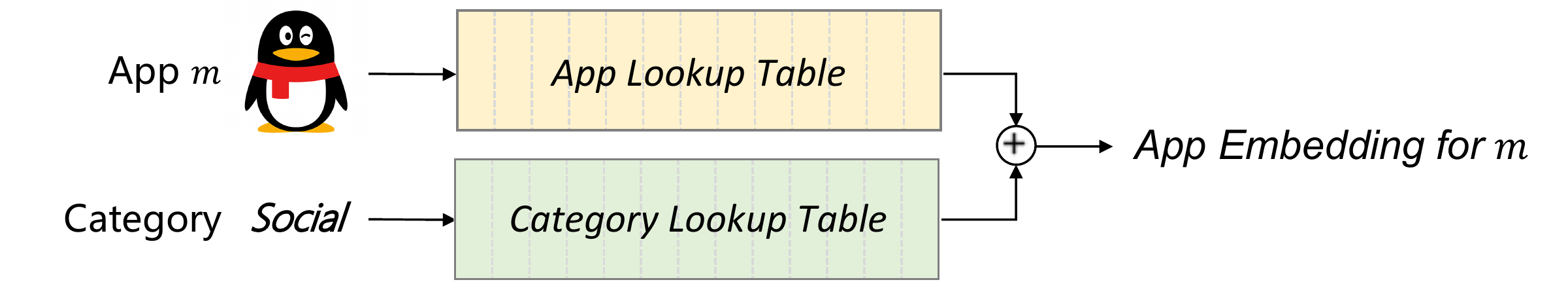}
  \caption{Illustration for App Embeddings}
  \Description{App Embeddings}
  \label{fig:app_embedding}
\end{figure}

We carefully design our weight matrices for several parts of the model, which helps to solve the sparsity problem and tightly couple the autoencoder part and the transformer parts. As shown in Figure~\ref{fig:app_embedding}, the app embeddings are built based on both the app ID and its corresponding category ID. Even if the usage of some app is gravely sparse, its category can still provide valid information. This setting helps to overcome the problem of sparsity.

As introduced previously, we repeatedly use the $M \times d_{model}$ embedding matrix for apps, \ie, at the input and output of the retention autoencoder, the input of the transformer encoder, the output for the masked app prediction, the output of the transformer decoder, as well as the reconstruction part for retention from the user embeddings~(bottleneck). We tie the weight matrices of all these parts together, \ie,
\begin{align}
  \mathbf{W}^{\Omega}=\mathbf{W}^{\Theta}=\mathbf{W}^{\Phi}=\mathbf{W}^{(4)}={\mathbf{W}^{a}}^{\mathrm{T}}.
\end{align}
We reduce the total number of parameters by tying weight matrices of the above layers, which benefits of overcoming the problem of sparsity. Moreover, weight tying benefits the backpropagation of the gradient and speeds the convergence.

\subsection{Modified Multi-head Self-attention}
\label{sec:self-attention}

\begin{figure}[ht]
  \centering
  \includegraphics[width=0.875\linewidth]{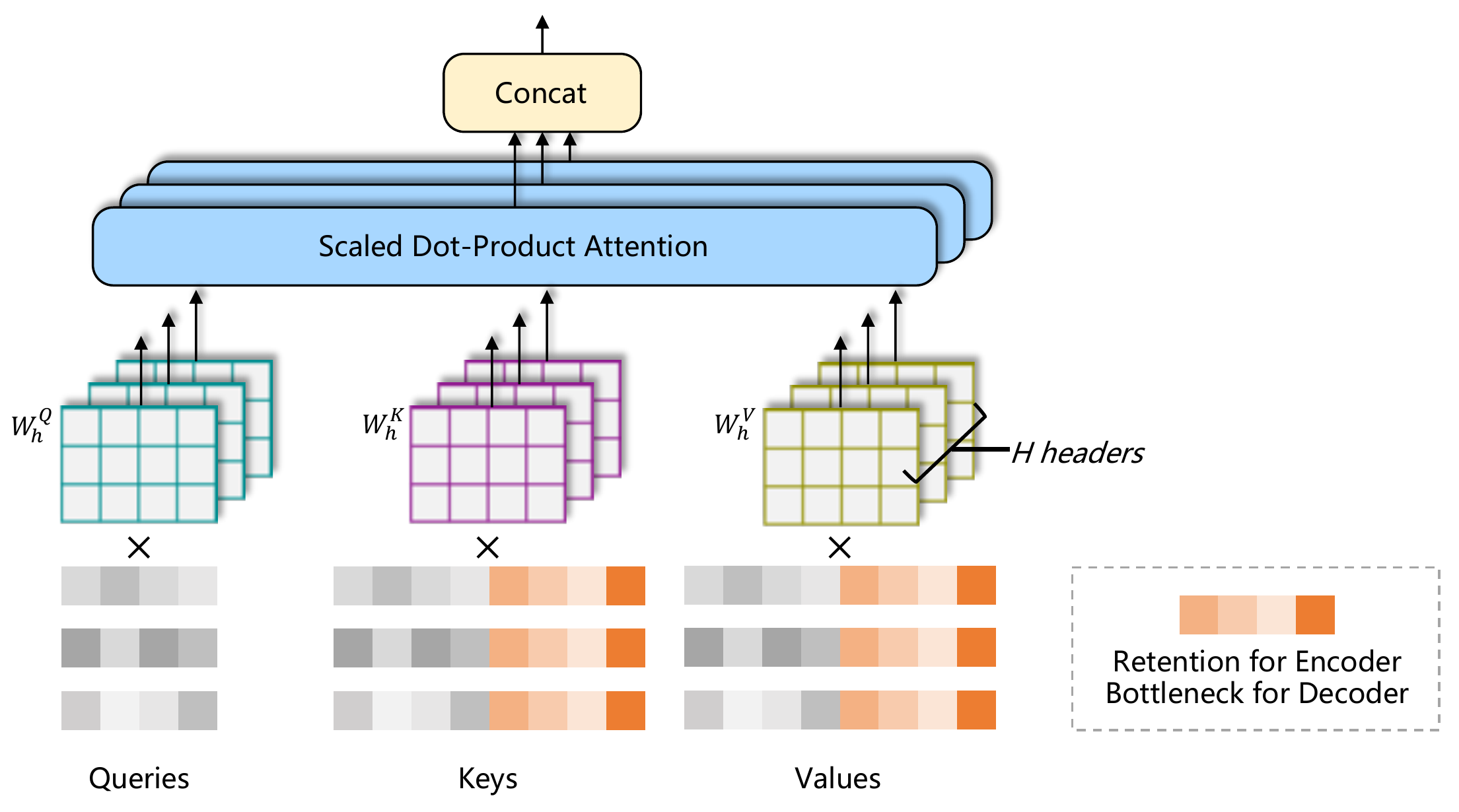}
  \caption{Modified multi-head self-attention. It is applied in both the transformer encoder and decoder.}
  \Description{Modified Multi-Head Self-Attention.}
  \label{fig:attention}
\end{figure}

In our scenario, retention, bottleneck~(user embeddings), installation, and uninstallation are heterogeneous. Each installation or uninstallation represents a single operation, but the retention or bottleneck is a cumulation of all the installation and uninstallation operations. Therefore, to better encourage the information interaction among retention, bottleneck, and (un)installation, the multi-head self-attention is modified, as shown in Figure~\ref{fig:attention}.

By concatenating the retention~(for the transformer encoder part) or bottleneck~(for the transformer decoder part) to each key and value for the scaled dot-product attention, we enforce the information interaction with retention or bottleneck in every attention step. In this way, the transformer encoder fuses the information from retention and (un)installation more efficiently, and the decoder extracts information from the bottleneck better for reconstruction tasks. This modification improves the quality of user embeddings, as shown by the experimental results.

\subsection{Multi-objective Joint Training Scheme}
\label{sec:train_scheme}

For model training, we apply a joint training scheme consisting of three tasks, \ie,

{\bfseries Task \#1: Main Reconstruction.} To generate general-purpose user embeddings on basis of their behaviors on mobile app usage, we train the proposed model to reconstruct all the retention, installation, and uninstallation information from the user embeddings. This task is indispensable in the joint training scheme and can be divided into two sub-tasks: (1)~Reconstructing the retention data from the user embeddings~(bottleneck layer) by a multi-layer perceptron network. We choose the \emph{sigmoid} cross-entropy as the loss function. (2)~Reconstructing the installation and uninstallation sequences by the transformer decoder. We calculate the loss of this sub-task by averaging the \emph{softmax} cross-entropy loss of every (un)installation. The sum of the losses of these two sub-tasks is the loss of this main reconstruction task, and we denote the loss as $\mathcal{L}_{main}$.

{\bfseries Task \#2: Auxiliary Retention Reconstruction.} This auxiliary task is for the autoencoder part. We also choose the \emph{sigmoid} cross-entropy as the loss function denoted as $\mathcal{L}_{aux}$. 

{\bfseries Task \#3: Masked App Prediction.} This task is similar to the ``Masked LM'' task in BERT~\cite{devlin2019bert}. We randomly mask apps in installation and uninstallation but keep the corresponding date and behavior type. The transformer encoder is trained only to predict the masked apps. For simplicity, we just follow the masking rate in BERT and abandon the ``random replacement or keep''. We calculate the loss of this task, denoted as~$\mathcal{L}_{mask}$, by averaging the \emph{softmax} cross-entropy loss of every masked app.

The final loss function of our model is the sum loss of above three tasks as well as the regularization loss, \ie, $\mathcal{L} = \mathcal{L}_{main} + \mathcal{L}_{aux} + \mathcal{L}_{mask} + \mathcal{L}_{reg}$. And $\mathcal{L}_{reg}$ is the $\ell 2$ norm regularization loss for all the trainable parameters.

\section{Offline Experiments}
\label{sec:experiment}
In this section, we demonstrate the offline performance of AETN in generating general-purpose user embeddings. We compare the baseline with four different versions of AETN in three typical downstream offline experiments. Then we show that the auxiliary retention reconstruction task for the autoencoder part can help the convergency of the transformer parts. Finally, we compare the user embeddings generated by the baseline and AETN intuitively.

\subsection{Dataset}
\label{sec:train_dataset}

We use real industrial data from Tencent for model training. Following the rules introduced in Section~\ref{sec:preprocessing}, we consider more than 10 thousand apps. Then we sample 20 million users and 500 million records of installation and uninstallation dated from 2019.07 to 2019.12. We randomly split out about 5 million users for validation.

\begin{table*}
  \caption{Offline evaluation results for user embeddings.}
  \label{tab:offline_results}
  \begin{tabular}{cccccccc}
    \toprule
    \multirow{2}{*}{Model} & \multicolumn{5}{c}{Next Week's Installation Prediction} & \multirow{2}{*}{\shortstack{Look-alike\\ Audience Extension}} & \multirow{2}{*}{\shortstack{Feed \\ Recommendation}}\\
    & Category \#1 & Category \#2 & Category \#3 & Category \#4 & Average\\
    \midrule
    DAE & 0.7294 & 0.7297 & 0.7844 & 0.7132 & 0.7392 & 0.8175 & 0.6358\\
    AETN w/o $\mathcal{L}_{mask}$ & 0.7903 & 0.7818 & 0.8166 & 0.7743 & 0.7908 & 0.8290 & 0.6395\\ 
    AETN w/o $\mathcal{L}_{aux}$ & 0.8024 & 0.7913 & 0.8196 & 0.7866 & 0.8000 & 0.8301 & 0.6403\\
    V-AETN & 0.8014 & 0.7924 & 0.8133 & 0.7746 & 0.7954 & 0.8307 & 0.6401\\
    AETN & \textbf{0.8026} & \textbf{0.7974} & \textbf{0.8215} & \textbf{0.7879} & \textbf{0.8023} & \textbf{0.8309} & \textbf{0.6406}\\
    \bottomrule
  \end{tabular}
\end{table*}

\subsection{Models}

We train and evaluate 5 models, including a baseline and 4 different versions of AETN, as follows.
\begin{itemize}
    \item \textbf{DAE}. Denoising autoencoder~\cite{vincent2008extracting,vincent2010stacked} is widely applied for unsupervised representation learning. We train it to generate user embeddings based on user retention data.
    \item \textbf{AETN w/o $\bm{\mathcal{L}_{mask}}$}. A degenerated version of AETN trained without the task of masked app prediction.
    \item \textbf{AETN w/o $\bm{\mathcal{L}_{aux}}$}. Another degenerated version of AETN trained without the auxiliary retention reconstruction task.
    \item \textbf{V-AETN}. The AETN with vanilla multi-head self-attention proposed in~\cite{vaswani2017attention}.
    \item \textbf{AETN}. The complete version of the model which is introduced in Section~\ref{sec:AETN}.
\end{itemize}

Details of model settings and hyper-parameter configurations are listed in Appendix~\ref{appendix:hyperparameter}.
RNN-based models are not involved. In addition to the uneven distribution of user behaviors, the low efficiency of the training makes them infeasible in our scenario.

\subsection{Offline Evaluation Tests}

We conduct our offline experiments on three typical downstream applications, including applications from both related domains and a different domain. The evaluation tasks are as follows:

{\bfseries Test \#1: Next Week's Installation Prediction}. This task is to predict which users are going to install specific (niche) categories of apps in next week. We collect data from about 5 million users and then divide them into a training set, a validation set, and a testing set in a 3:1:1 ratio. After generating user embeddings, we train multi-layer perceptron networks to predict whether one would install apps of four categories in next week.

{\bfseries Test \#2: Look-alike Audience Extension}. This is a common task in computational advertising~\cite{zhang2016implicit,mangalampalli2011feature}. We use a dataset containing about half a million users with about 10\% seed users for an out-of-vocabulary niche app. Following the common practice, we train XGBoost~\cite{chen2016xgboost} look-alike models to evaluate different user embeddings, and report the 10-fold cross-validation results.

{\bfseries Test \#3: Feed Recommendation}. To evaluate the universal user embeddings in a cross-domain scenario, we use feed recommendation data from the ``Find'' tab of \emph{Tencent Wi-Fi manager}. We select about 1.2 million users and extract their behaviors in 8 days, then use the data from the first 7 days for training and the data from the last day for validation and testing. The training set contains about 27 million records, and the validation set and the testing set contain approximately 2 million records, respectively. We train Deep \& Cross Networks~\cite{wang2017deep} based on the generated user embeddings as well as other features for feed recommendations.

In all three tasks, we use the area under the ROC curve~(AUC) as the metric. We run each test 5 times and report the average.

\subsection{Offline Evaluation Results}
\label{sec:result}
Table~\ref{tab:offline_results} shows the results in all three downstream experiments. We can draw the following conclusions from the results.

{\bfseries All versions of AETN perform better than DAE.} In next week's installation prediction, AETN brings an average AUC improvement by 0.0631 for all four categories. The rest two applications still enjoy improvements by 0.0134 and 0.0048. This is a significant improvement for industrial applications where 0.1\% AUC gain is remarkable~\cite{ma2018entire}. Such improvement confirms two hypotheses. Firstly, short-term user interests contained in installation and uninstallation are valuable for various downstream applications, to different extent. Secondly, the proposed AETN is capable of extracting long-term and short-term user interests from all types of user behaviors and compressing them together into the user embeddings.

{\bfseries The masked app prediction task makes an important contribution to improve the quality of user embeddings.} It brings up an average AUC by 0.0115 in next week's installation prediction. Even for the look-alike audience extension and the feed recommendation, the AUC lift brought by this task is over 0.0010. We owe this to that the masked app prediction not only helps the transformer encoder extract information more efficiently, but also brings a data augmentation effect in the training process.

{\bfseries The modified multi-head self-attention performs better than the vanilla one.} The simple modification, which encourages information interaction among retention, bottleneck, and (un)installation, contributes an AUC gain of 0.0069 to the next week's installation prediction.

{\bfseries The auxiliary retention reconstruction also benefits the quality of generated user embeddings.} Without this auxiliary task for the autoencoder part, the AUC in the next week's installation prediction goes down by 0.0023. Besides the improvement in user embeddings, we find that the training efficiency is also improved by this auxiliary retention reconstruction.

\subsection{Training Efficiency Comparison}
\begin{figure}[!t]
\centering
\subfloat[]{\includegraphics[width=0.49\linewidth]{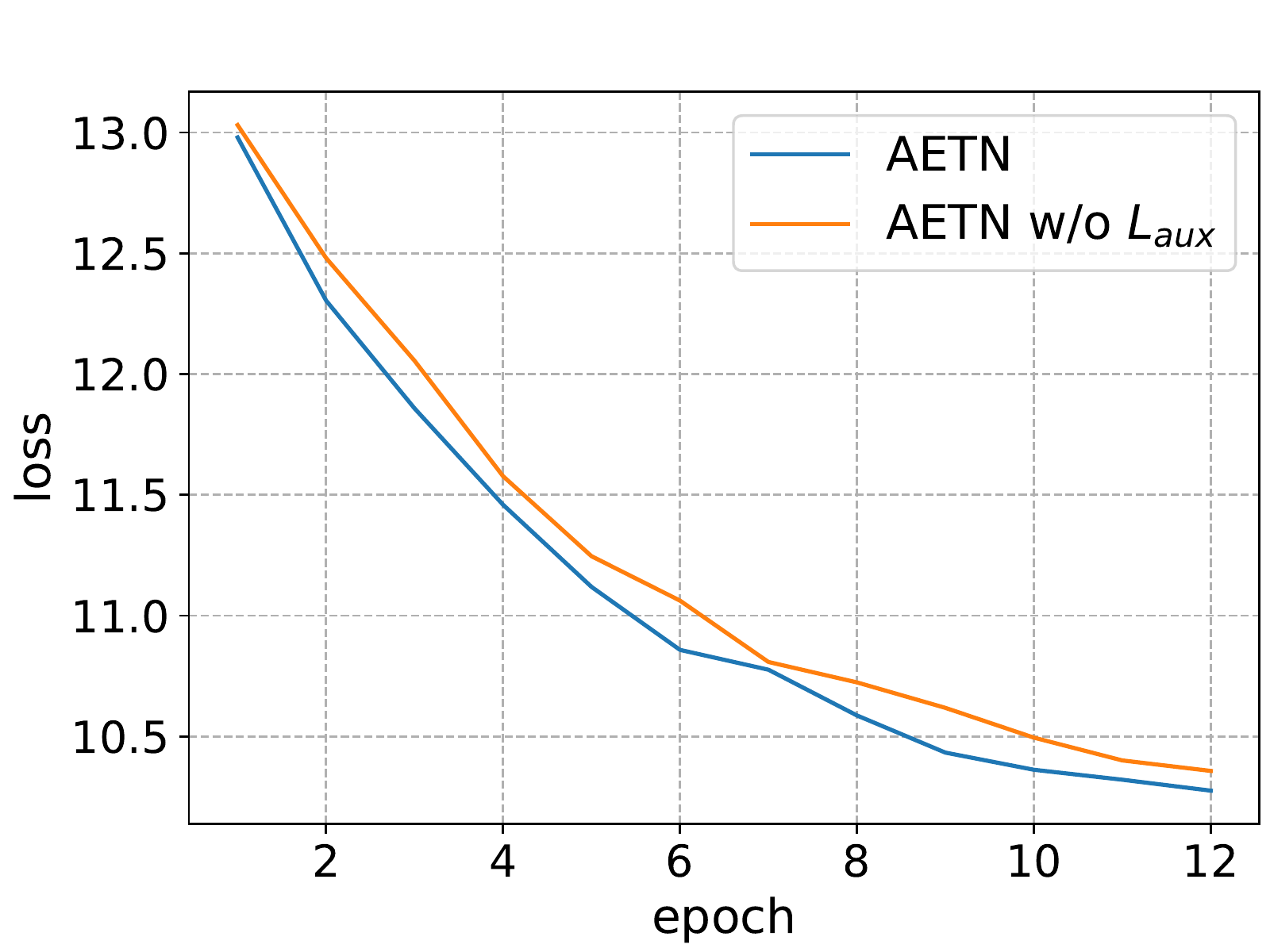}
\label{fig:reduce_loss}}
\hfil
\subfloat[]{\includegraphics[width=0.49\linewidth]{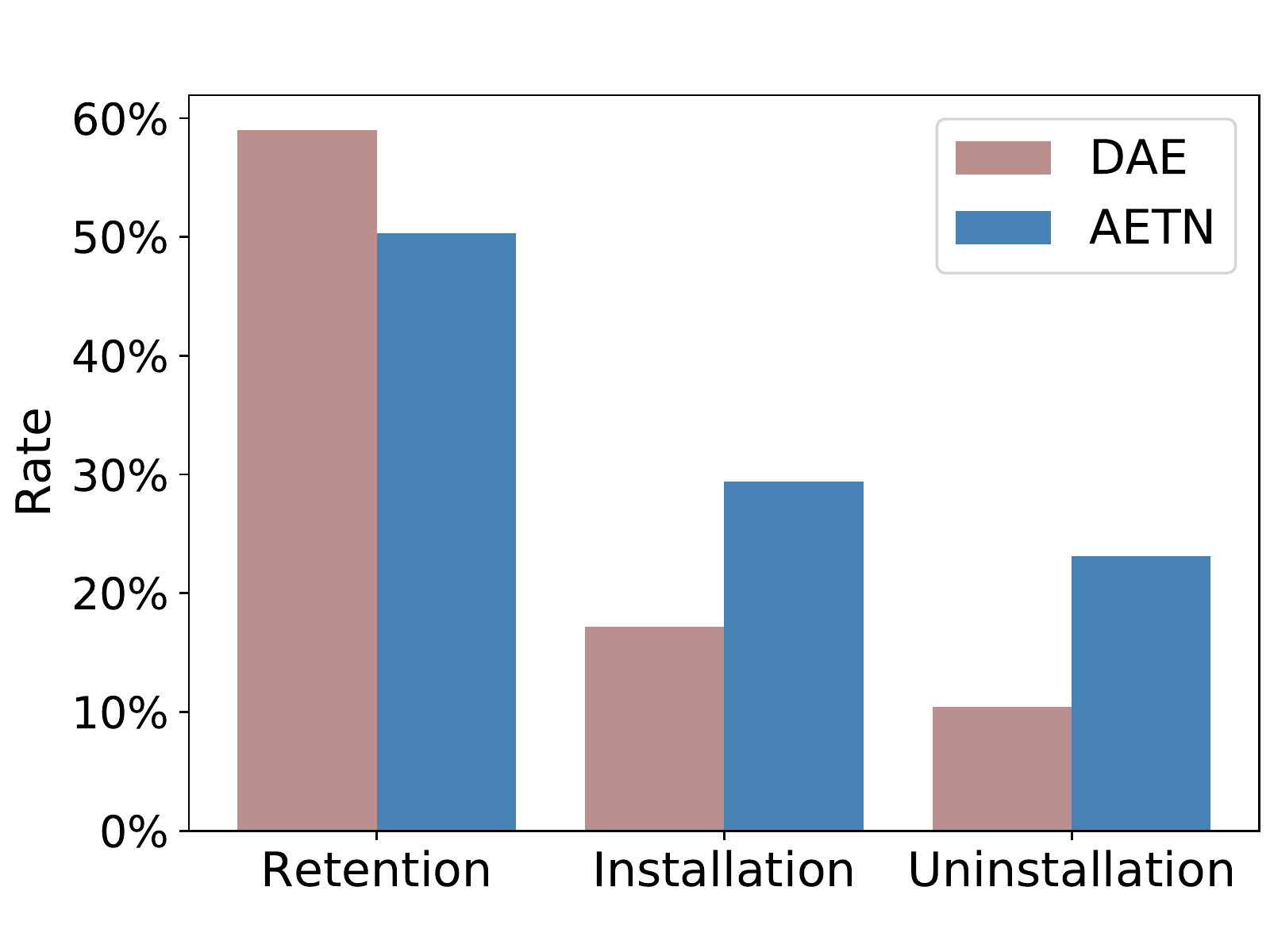}
\label{fig:neighbor}}
\caption{(a)~Records for the sum of $\mathcal{L}_{main}$ and $\mathcal{L}_{mask}$ on validation dataset. Two models are trained with the same settings except for the auxiliary retention reconstruction. (b)~The overlap rate of apps between 10 thousand pairs of neighbor users.}
\end{figure}

When training AETN and AETN w/o $\mathcal{L}_{aux}$, we monitor the sum of $\mathcal{L}_{main}$ and $\mathcal{L}_{mask}$ on the validation dataset, to confirm the improvement in training efficiency brought by the auxiliary retention reconstruction. As shown in Figure~\ref{fig:reduce_loss}, the auxiliary task makes the transformer parts in the AETN converge faster. With the autoencoder and weight tying, gradients from the output layer can be passed through fewer layers to the app embedding matrix. Moreover, the complete version of AETN also achieves a lower loss when both models eventually converge.

\subsection{App Overlap between Neighbor Users}

To intuitively compare the output embeddings by AETN and DAE, we measure the overlap rate of apps between pairs of neighbor users based on the embeddings. For each user, we choose the corresponding user with the least Euclidean distance according to the embeddings as the neighbor. We randomly sample 10 thousand users and find their neighbors from 1 million randomly-selected users. For each pair of neighbors, we calculate the overlap rate of apps in retention, installation, and uninstallation. Figure~\ref{fig:neighbor} shows the average results of all the neighbor pairs for both the AETN embeddings and DAE embeddings. The results show that the AETN succeeds in injecting information from installation and uninstallation into user embeddings and maintaining the majority of the retention. At the same time, the DAE embeddings, which we extract only based on the retention information, cannot provide much information regarding the installation and uninstallation.

\section{Online A/B Testing}
\label{sec:online}

To further verify the effectiveness of the output user embeddings, we conduct online feed recommendation A/B testing from 2020-02-01 to 2020-02-10, in the ``Good Morning'' tab of \emph{Tencent Mobile Manager} and the ``Find'' tab of \emph{Tencent Wi-Fi Manager}. We split online A/B test traffic by userIDs evenly for the tested models. We evaluate the base models, models with DAE embeddings, and models with AETN embeddings. The improvement results compared with the base models are reported in Table~\ref{tab:online_results}.

\begin{table}
  \caption{Online evaluation results for models with DAE embeddings and models with AETN embeddings.}
  \label{tab:online_results}
  \resizebox{\linewidth}{!}{
  \begin{tabular}{cccccc}
    \toprule
    Tab & Model & UV CTR & PV CTR & Engagement & Clicks\\
    \midrule
    \multirow{2}{*}{\shortstack{Good \\ Morning}} & +DAE & +1.26\% & +2.23\% & +2.27\% & +2.11\% \\
     & +AETN & +2.81\% & +4.79\% & +5.24\% & +3.86\% \\
    \midrule
    \multirow{2}{*}{Find} & +DAE & +4.12\% & +2.29\% & +2.82\% & +2.11\%\\
     & +AETN & +4.15\% & +5.96\% & +8.14\% & +3.23\% \\
    \bottomrule
  \end{tabular}
  }
\end{table}

We mainly consider the following metrics. \emph{UV CTR} measures the click-through rate in terms of the user view, and \emph{PV CTR} measures the click-through rate in terms of the page view. \emph{Engagement} measures the average staying time of each user. \emph{Clicks} measures the average number of articles each user reads. From the table, we can find that compared with the base models, all considered metrics enjoy improvements by AETN embeddings, ranging from 2\% to 8\%. Compared with DAE embeddings, \emph{PV CTR}, and \emph{Engagement} enjoy more substantial improvements brought by the AETN embeddings, and we hypothesize that AETN introduces the installation and uninstallation information, thus could capture short-term interests of users in addition to the long-term interest from retention, and this information is more critical to \emph{PV CTR} and \emph{Engagement}. Comparing the results of the ``Good Morning'' tab and the ``Find'' tab, we can find that the improvements in the ``Find'' tab are more significant. It may be due to that users tend to read articles in the ``Find'' tab all along the day, in contrast to the ``Good Morning'' tab where users majorly read the news after getting up in the morning. The exposure per user in the ``Find'' tab is significantly more. Therefore, better modeling for user interests is even more critical.

\section{Deployment}
\label{sec:deploy}

We implement the model with Tensorflow~\cite{abadi2016tensorflow}. It takes about 60 hours for training using 4 NVIDIA Tesla M40 GPUs. As the embeddings represent both long-term and short-term interests of users, it is crucial to keep updating the embeddings for the best performance. However, a large number of users bring challenges to update frequently. Generally, we have two strategies for updating:

\begin{itemize}
    \item \textbf{Model Updating}. We may update the model for the best performance. This method takes into consideration the emerging apps with an utterly up-to-date app list and the distribution of the data. However, updating the model changes the semantic structure of user embeddings completely. Thus, we need to update all downstream models simultaneously.
    \item \textbf{Feature Updating}. We can also keep the model fixed and only update the features of users. Thus we have the up-to-date behaviors of users taken into consideration, and the updated embeddings can still be in the same semantic space. This strategy makes the updating less expensive.
\end{itemize}

In practice, we find that feature updating is more cost-effective for downstream applications, which is because the apps usually do not change drastically within a few months. However, updating the embeddings for billion-scale users is still challenging. To reduce computation, we only update the representations of active users of downstream applications every day. This strategy can reduce the number of users that need to be updated each time to the order of millions. The model could be updated much less frequently. Once the model is updated, we use a new feature ID to prevent confusion.

\section{Related Work}
\label{sec:relatedwork}

We summarize the related work in three fields, including applications with app behavior data, unsupervised feature extraction and transformer networks.

\subsection{Applications with App Usage Data}

User behaviors on mobile apps usage contain rich preference information and have been used in a variety of applications~\cite{lu2014mining}. The most significant of which is app install advertisements~\cite{globaladspend2018, lee2016targeting} and mobile app recommendations~\cite{zhu2014mobile}. Yahoo posted a large scale prediction engine for app install advertising based on a two-step logistic regression model considering user features generated from behaviors on apps~\cite{bhamidipati2017large}. For reducing sparseness, Yahoo also classifies apps into predefined interest taxonomies when understanding app usage patterns~\cite{radosavljevic2016smartphone}. Usage patterns of apps are learned for app purchase recommendations with a Deep Memory Network\cite{gligorijevic2018modeling}. Beyond app install advertising, users’ app-installation behaviors are also used for news recommendations~\cite{liu2017transfer}, where the knowledge of the neighborhood of the cold-start users is transferred from an APP domain to a new domain.
A large survey on mobile app user behaviors across main app markets around the world was conducted to instruct cross-country app competitions and analyze the challenges for software engineering~\cite{lim2014investigating}.

In this paper, we address the real-life need of general-purpose user embeddings based on user behaviors on app usage. The user embeddings can be used for a variety of downstream applications.

\subsection{Unsupervised Representation Learning}

Unsupervised representation learning is a long-standing problem~\cite{bengio2013representation,zhang2018network}, and autoencoders have been deployed successfully in many real-world applications~\cite{baldi2012autoencoders}. It follows an encoder-decoder structure and tries to reconstruct the input through a bottleneck layer. Sparse autoencoders~\cite{liu2016hsae}, denoising autoencoders~\cite{vincent2008extracting, vincent2010stacked}, variational autoencoders~\cite{pu2016variational}, 
adversarial autoencoders~\cite{makhzani2015adversarial},
and so on, have been proposed as extensions. Recently, more advanced unsupervised representation learning has been proposed, including BERT~\cite{devlin2019bert} for natural language processing, and MoCo~\cite{he2019momentum} for computer vision, 
With a large amount of data and deep models, unsupervised representation learning is able to achieve comparable or even better performance with fewer annotations than traditional supervised learning~\cite{devlin2019bert,he2019momentum}.

In this paper, an unsupervised representation learning from user behaviors on mobile apps is presented. We address the unique challenges of this problem with the tailored autoencoder-coupled transformer network, and demonstrate the effectiveness.

\subsection{Transformer Networks}

The transformer model was first introduced in \cite{vaswani2017attention}, and has been used widely for modeling sequences in natural language processing tasks~\cite{devlin2019bert}, recommendations~\cite{sun2019bert4rec,chen2019bert4sessrec},  and music generations~\cite{huang2018music}. Transformers can simultaneously attend to every token of their input sequence with self-attention mechanism, 
and it is proved that a multi-head self-attention layer with a sufficient number of heads is at least as expressive as any convolutional layer~\cite{cordonnier2019relationship}.
Compared with recurrent neural networks like long-short term memory~(LSTM)~\cite{hochreiter1997long}, transformers are more parallelizable and require significantly less time to train on large datasets~\cite{vaswani2017attention}.
Transformer-XL~\cite{dai2019transformer} and reformer~\cite{kitaev2020reformer} are proposed to further reduce the complexity when the length of sequences is very long, \eg, sequences of length 10,000.

In this paper, we couple a transformer network with an autoencoder to model the retention, installation, and uninstallation collectively. We modify
the vanilla transformer in order to emphasize the retention state or user embeddings when the installation and uninstallation are being modeled.

\section{conclusions}
\label{sec:conclusion}

In this paper, we present our recent practice for unsupervised user embedding learning based on mobile app usage. To address the unique challenges of this problem in the real system, we propose a tailored model called AutoEncoder-coupled Transformer Network~(AETN). Extensive online and offline experimental results demonstrate the effectiveness of the proposed model. We also introduce the details about the deployment. The output general-purpose user embeddings can fertilize multiple downstream applications that require user representations at Tencent. Now the output embeddings have been serving the feed recommendation scenes in \emph{Tencent Mobile Manager} and \emph{Tencent Wi-Fi Manager}. In the future, we plan to explore fine-tuning the transformer encoder part for learning task-specific user embeddings.

\begin{acks}
The authors would like to thank the BlueWhale project team of Tencent for supporting our research.
\end{acks}

\bibliographystyle{ACM-Reference-Format}
\bibliography{references}

\appendix

\section{Details of Hyper-parameters and Experiment Settings}
This appendix provides detailed supplementary information for the model settings, the hyper-parameter configurations, as well as the experiment settings.
Readers may refer to the publicly-available code for more implementation details.

\subsection{Model Settings and Hyper-parameter Configurations}
\label{appendix:hyperparameter}

There are many settings for the model and hyper-parameters for the training process. To balance the efficiency and performance, we directly determine some of them based on our previous experience and find the optimal settings for the others according to the main reconstruction loss on the validation dataset and the performance of the generated user embeddings on a downstream experiment.

In terms of the basic structure of the AETN, we choose to use an autoencoder with three hidden layers, as well as two transformer encoder layers and one transformer decoder layer. The dimension of the first hidden layer in the autoencoder and the hidden size of transformers, \ie, $d_{model}$, are set to $512$. We set the hidden size of the position-wise feed-forward networks in the transformers to $1024$, and the number of self-attention headers is 8. Considering the limitation of data storage, computation complex, and time delay in downstream applications, the size of the bottleneck layer~(\ie, the dimension of user embeddings) is $128$.

In terms of the regularization, the dropout rate of the input layer of user retention is set to $0.05$, and the dropout rate for multi-head self-attention mechanism and the position-wise feed-forward networks is set to $0.1$.

By monitoring the loss of the main reconstruct task on the validation dataset, we choose to use Adam as the optimizer and the mini-batch size is $1000$. We also choose to apply exponential decay, in which the learning rate starts at $0.0001$, and the decay rate is set to $0.8$ per epoch. For the $\ell 2$ norm regularization, we set the factor to $1.5\mathrm{e}{-7}$.

The length of the installation or uninstallation sequences is another important hyper-parameter that influences the quality of user embeddings. We train several models when setting the length to 15, 20, 25, 30, and 35. Then we generate and evaluate different versions of user embeddings in the feed recommendation test, and determine the optimal length as 25.

When training the proposed AETN with the task of masked app prediction, we follow the masking rate in BERT. Therefore, we mask 3 apps in installation sequences and another three in uninstallation sequences in the training process. Note that we only mask apps when training models, complete sequences for installation and uninstallation are kept for the validation and prediction.

The baseline model in the offline evaluation tests, DAE, shares the same structure with the autoencoder in AETN. The dimension of the bottleneck layer of DAE is also set to $128$.

\subsection{Additional Details for Offline Evaluation Datasets}
The four categories we selected in the next week's prediction is four typical niche ones that need app advertising to enlarge their user base. Apps from these categories are also the long-tailed ones suffer from serious sparsity. The average installation rates for these four categories are approximately $600$, $400$, $25$, and $300$ per million people, respectively.

\subsection{Measurement of App Overlap}
We present the detailed measurement of app overlap between a pair of neighbor users. For user $U$, we find her neighbor $V$ and get the retention, installation, and uninstallation of them. The overlap rate of apps in user retention is calculated by dividing the number of apps in $U$'s retention by the number of apps in the intersection of $U$'s and $V$'s retention. In terms of app overlap in installation or uninstallation, the dates of the operations, as well as the repetitive operations on one app, are not considered. Therefore, we firstly transform the apps sequences into app sets. Then the overlap rate of apps in (un)installation is calculated by dividing the size of $U$'s (un)installation set by the size of the intersection of $U$'s and $V$'s (un)installation set.

\end{document}